\documentclass[a4paper,conference]{IEEEtran}

\usepackage{setspace}
\usepackage{times}
\usepackage{epsfig}
\usepackage{graphicx}
\usepackage{amsmath}
\usepackage{amssymb}
\usepackage{pifont}
\usepackage{color}
\newcommand{\tabincell}[2]{\begin{tabular}{@{}#1@{}}#2\end{tabular}}
\usepackage[numbers]{natbib}
\usepackage{multirow}
\usepackage{comment}
\usepackage[misc]{ifsym}
\usepackage[pagebackref=true,breaklinks=true,letterpaper=true,colorlinks,bookmarks=True]{hyperref}

\renewcommand\thesection{\arabic{section}}

\newcommand{\yq}[1]{\textcolor{black}{#1}}
\newcommand{\why}[1]{\textcolor{black}{#1}}

\begin{document}

\title{The First Competition on Resource-Limited Infrared Small Target Detection Challenge: Methods and Results}


\author{Boyang Li$^{*}$, Xinyi Ying$^{*}$, Ruojing Li$^{*}$, Yongxian Liu$^{*}$, Yangsi Shi$^{*}$, Miao Li$^{\textrm{\Letter}}$, Xin Zhang$^{*}$, Mingyuan Hu$^{*}$\\
Chenyang Wu$^{*}$, Yukai Zhang$^{*}$, Hui Wei$^{*}$, Dongli Tang$^{*}$,  Jian Zhao$^{*}$, Lei Jin$^{*}$,  Chao Xiao$^{*}$, Qiang Ling$^{*}$\\
Zaiping Lin$^{*}$, Weidong Sheng$^{*}$, Chenxu Peng, Wei Ke, Yuhang He, Huoren Yang, Lingjie Liu, Ziqi Liu\\
Zelin Shi, Yunpeng Liu, Chuang Yu, Jinmiao Zhao, Heng Xiang, Shan Yang, You Li, Guoqing Wang\\
Tianyu Li, Minghang Zhou, Dongyu Xie, Chenxi Lan, Chaofan Qiao, Yupeng Gao, Yongxu Liu, Wei Feng \\
Zhihao Ma, Deping Chen, Xiaopeng Song, Jiuping Yang, Zhaobing Qiu, Rixiang Ni, Ye Lin, Changhai Luo\\
Shuyuan Zheng, Baojin Huang, Xiaoqi Zhou, Qingshan Guo, Xuyang Zhang,
Xiuhong Li, Dangxuan Wu\\
Jian Ma, Haodong Zeng, Luyao Wang, Qiang Fu, Yimian Dai, Renke Kou, Jian Song, Changfeng Feng\\
Mengxuan Xiao, Zihao Xiong, Hong Huang, Yingxu Liu, Quanyi Zhao
}


\maketitle
\section{Abstract}
In this paper, we briefly summarize the first competition on resource-limited infrared small target detection (namely, LimitIRSTD). This competition has two tracks, including weakly-supervised infrared small target detection (Track 1) and lightweight infrared small target detection (Track 2). 46 and 60 teams successfully registered and took part in Tracks 1 and Track 2, respectively. The top-performing methods and their results in each track are described with details. This competition inspires the community to explore the tough problems in the application of infrared small target detection, and ultimately promote the deployment of this technology under limited resource.

\footnotetext{
	\noindent $^{*}$Boyang Li, Xinyi Ying, Ruojing Li, Yongxian Liu, Yangsi Shi, Miao Li, Xin Zhang, Mingyuan Hu,
	Chenyang Wu, Yukai Zhang, Hui Wei, Dongli Tang,  Jian Zhao, Lei Jin,  Chao Xiao, Qiang Ling, Zaiping Lin, Weidong Sheng are the ICPR-2024 LimitIRSTD challenge organizers, while the other authors participated in this challenge.	\\
	\noindent $^{\textrm{\Letter}}$Corresponding author: Miao Li\\
	~~Section \ref{appendix} provides the authors and affiliations of each team.\\
	~~ICPR 2024 webpage: \url{https://icpr2024.org/}\\
	~~Challenge webpage: \url{https://limitirstd.github.io/}\\
	~~Leaderboard Track1: \url{https://bohrium.dp.tech/competitions/8821868197?tab=introduce}\\
    ~~Leaderboard Track2: \url{https://bohrium.dp.tech/competitions/9012970343?tab=introduce}\\
	~~BasicIRSTD toolbox: \url{https://github.com/XinyiYing/BasicIRSTD}}

\IEEEpeerreviewmaketitle
\section{Introduction}

Infrared small target detection is widely used in many applications, such as marine resource utilization\cite{li2024mixed,2022light}, high-precision navigation\cite{liu2021nonconvex,wu2023mtu}, and ecological environment monitoring\cite{li2023direction,liu2023infrared}. Compared to generic object detection, infrared small target detection has several unique characteristics:
1) \textbf{Small:} Due to the long imaging distance, infrared targets are generally small, ranging from one pixel to tens of pixels in the images. 2) \textbf{Dim:} Infrared targets usually have low signal-to-clutter ratio (SCR) and are easily immersed in heavy noise and clutter background. 3) \textbf{Shapeless:} Infrared small targets have limited shape characteristics. 4) \textbf{Changeable:} The sizes and shapes of infrared targets vary a lot among different scenarios.

In recent years, remarkable progress in infrared small target detection has been witnessed with deep learning techniques. However, most approaches focus on fixed image resolution, single wavelength, limited imaging system, fully-supervised annotation, and oversized model. How to achieve wide area detection on multiple image resolution (e.g., 256, 512, 1024), wavelength (e.g., shortwave infrared, near-infrared, and thermal) at varied imaging systems (e.g., land-based, aerial-based, and space-based imaging systems) with weaker supervision and more lightweight model remains challenging.

The first LimitIRSTD competition is built upon our recently released WideIRSTD datasets, including  WideIRSTD-Full and WideIRSTD-Weak, for fully-supervised and single-point supervised small target detection, respectively. Specifically,  we focus on two competition tracks: 

\begin{itemize}
	\item{Track 1:  Weakly Supervised Infrared Small Target Detection Under Single Point Supervision.}  
	\item{Track 2:  Lightweight Infrared Small Target Detection Under Pixel-level Supervision.} 
\end{itemize}

 The best-performing methods and their results in each track are summarized with analyses in this paper.

\section{The First LimitIRSTD Challenge}
\subsection{Dataset}

\begin{figure}
	\centering
	\includegraphics[width=0.45\textwidth]{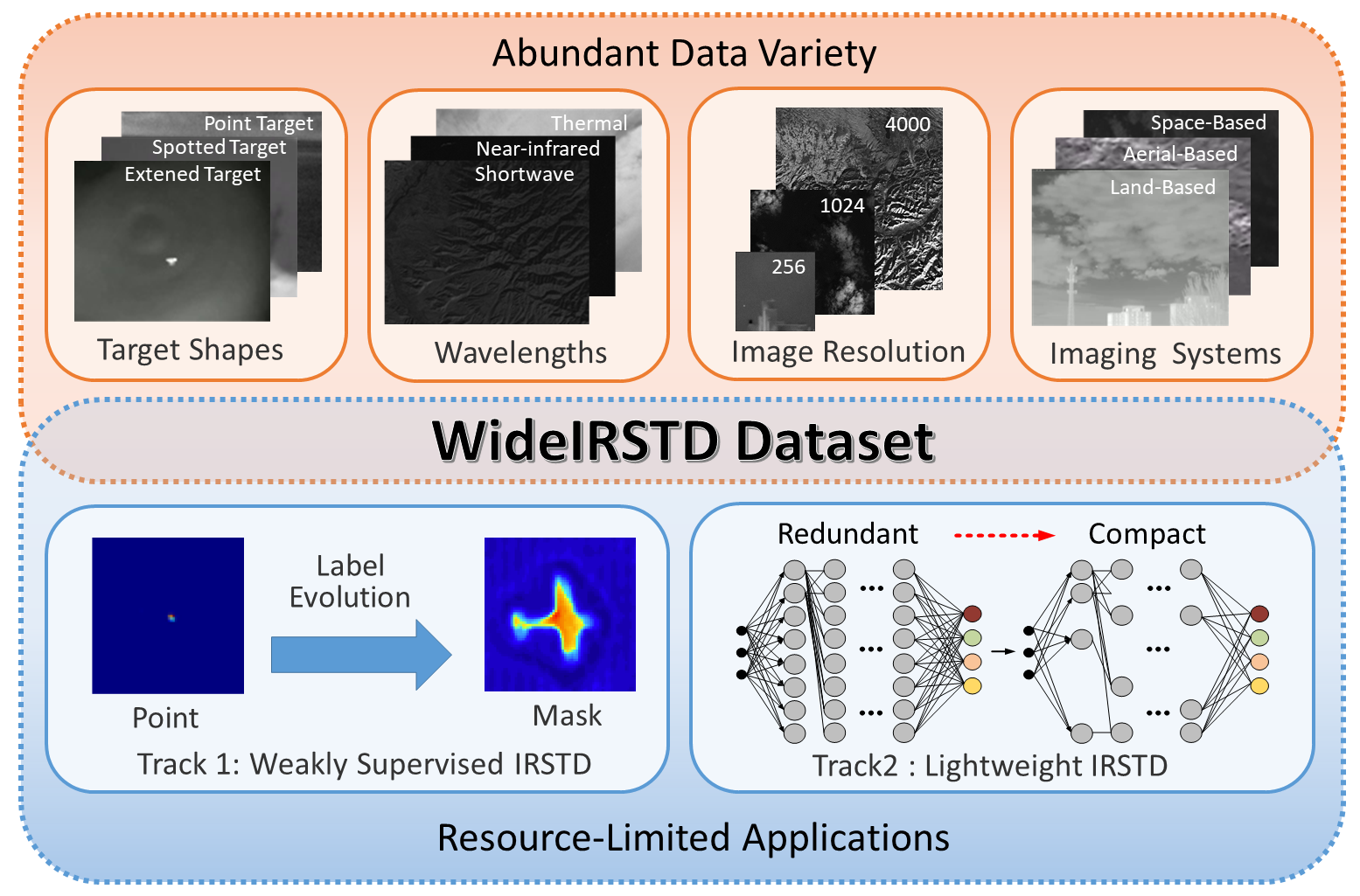}
	\caption{An illustration of WideIRSTD dataset.}
	\label{Fig-dataset}
\end{figure}

\begin{figure}
	\centering
	\includegraphics[width=0.45\textwidth]{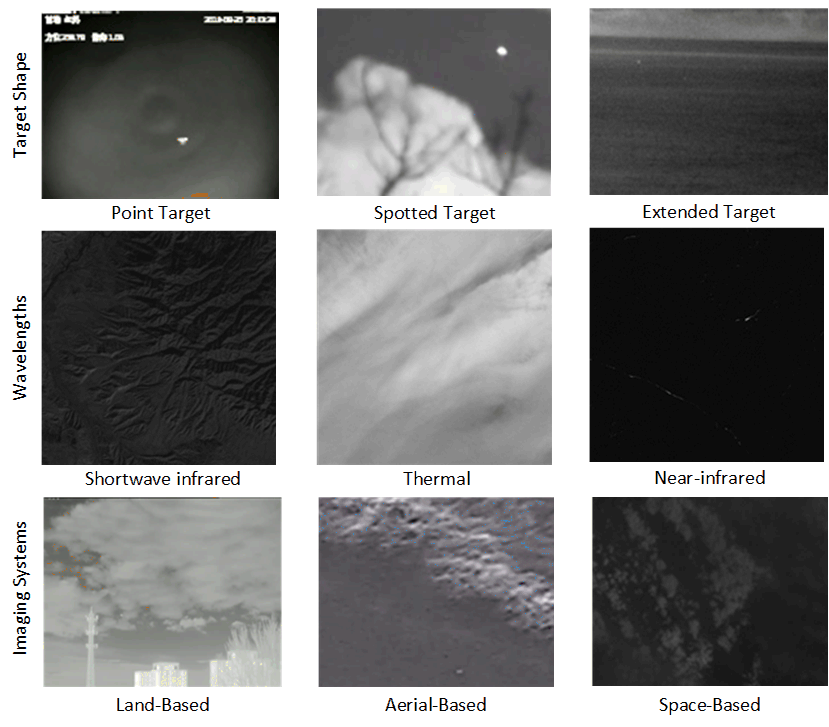}
	\caption{An illustration of WideIRSTD dataset.}
	\label{Fig-example}
\end{figure}

WideIRSTD\footnote{\url{https://github.com/XinyiYing/WideIRSTD-Dataset}} dataset consists of seven public datasets SIRST-V2\cite{NUAA-SIRST}, IRSTD-1K \cite{ISNet}, IRDST \cite{IRDST}, NUDT-SIRST \cite{DNANet} NUDT-SIRST-Sea \cite{NUDT-SIRST-Sea}, NUDT-MIRSDT \cite{DTUM}, Anti-UAV \cite{Anti-UAV} and a extra dataset developed by the competition (i.e., National University of Defense Technology), including simulated land-based and space-based data, and real manually annotated space-based data. As shown in Fig.~\ref{Fig-dataset}, the dataset contains images with various target shapes (e.g., point target, spotted target, and extended target), wavelengths (e.g., near-infrared, shortwave infrared and thermal), image resolution (e.g., 256, 512, 1024, and 3200, etc.), at varied imaging systems (e.g., land-based, aerial-based, and space-based imaging systems). Fig.~\ref{Fig-example} shows some example images of the WideIRSTD dataset.

In this challenge, this dataset is used to evaluate the performance of infrared small target detection (IRSTD) under resource-limited conditions (i.e., Track 1: Weakly Supervised IRSTD Under Single Point Supervision, Track 2: Lightweight IRSTD Pixel-level Supervision). 

For Track 1,  WideIRSTD-Full dataset is used. 6000 images with coarse point annotation (i.e., GT point is located around the centroid of the GT mask under Gaussian distribution) are used for training, and 500 images are used for test. For Track 2,  WideIRSTD-Weak dataset is used. 9000 images with groundtruth (GT) mask annotations are used for training, and 2000 images are used for test.

\subsection{Tracks}

\smallskip\noindent\textbf{Track 1:  Weakly Supervised Infrared Small Target Detection Under Single Point Supervision.} The goal of this task is to achieve weakly supervised infrared small target detection under single coarse point supervision (i.e., GT point is located around the centroid of the GT mask under Gaussian distribution).

\smallskip\noindent\textbf{Track 2: Lightweight Infrared Small Target Detection Under Pixel-level Supervision.} This task aims at achieving lightweight and efficient infrared small target detection under pixel-level supervision (i.e., the model achieves better detection performance with fewer parameters, less computation, and less memory usage). 

\subsection{Evaluation Metrics}
\label{sec:5}
The \why{performance} of each submitted result is automatically evaluated by the organizers using the following evaluation metrics.

 Pixel-level metric (i.e., intersection over union $IoU$) and target-level metrics (i.e., probability of detection $P_d$ and false-alarm rate $F_a$) are used as metrics for performance evaluation in Track 1. Note that, the
 linear weighted summation of $IoU$ and $P_d$ is used for ranking, and the corresponding performance score $S_p$ is defined as:
 \begin{equation}\label{eq1}
 	S_p=\alpha\times IoU+(1-\alpha)\times P_d,
 \end{equation} where $\alpha$ represents the weight. For Track 2, besides the performance score, we also introduce the number of parameters $P$ and $FLOPs$ for performance evaluation. Note that, we rank entries based on the sum of these three values, with each value normalized by a baseline state-of-the-art model for the task. The corresponding efficiency score $S_e$ is defined as:
 \begin{equation}\label{eq2}
 	S_e=1-\frac{P_{sub}/P_{base}+F_{sub}/F_{base}}{2},
 \end{equation} where $[\cdot]_{sub}$ and $[\cdot]_{base}$ represent the corresponding values of submission and baseline model. The final score $S_{pe}$ is the linear weighted summation of the performance score and the efficiency score, and can be defined as:
 \begin{equation}\label{eq3}
 	S_{pe}=\beta\times S_p+(1-\beta)\times S_e,
 \end{equation} where $\beta$ represents the weight.

\section{Competition Results}\label{sec4}
    In each track, we asked the top-performing teams to submit their codes and factsheets. In this section, we briefly describe their results and rankings. 

	  \begin{table*}
		\renewcommand\arraystretch{1.5}
		\caption{The award-winning results in the track of weakly-supervised infrared small target detection.}
		\label{tab1}
		\centering
		\scriptsize
		\setlength{\tabcolsep}{1.2mm}{
			\begin{tabular}{cllccccc}
				\hline
				Rank &Team
				& Authors
				& Score
				& mIoU ($\times 10^{-2}$)
				& $P_d$ ($\%$)
				& $F_a$ ($\times 10^{-6}$)
				
				\tabularnewline
				\hline
				1 & Chainey & \tabincell{l}{Chenxu Peng} & 60.7869 & 45.3729 & 76.2010  & 24.8629
				\tabularnewline
				2 & XJTU-IR & \tabincell{l}{Yang et al.} & 60.3803 & 42.5640 & 78.1966 & 26.5012
				\tabularnewline 
				3 & MCV-TEAM & \tabincell{l}{Yu et al.} & 60.0276 & 38.9762 & 81.0790 & 24.6809
				\tabularnewline
				4 & Stars Twinkle and Shine & \tabincell{l}{Xiang et al.} & 55.2378 & 42.4787 & 67.9970 & 12.4483
				\tabularnewline
				5 & MIG & \tabincell{l}{Zhou et al.} & 54.1322  & 35.9067 & 72.3577  & 38.687
				\tabularnewline
				6 & ISTDGroup-XDH & \tabincell{l}{Ma et al.} & 53.8369 & 39.8986 & 67.7753 & 10.1781
				\tabularnewline
				7 & Banyan City & \tabincell{l}{Ni et al.} & 53.645 & 34.5628 & 72.7272 & 15.5559
				\tabularnewline
				8 & GrapeJasmine & \tabincell{l}{Shen et al.} & 52.5642 & 38.8313 & 66.2971 & 6.5323
				\tabularnewline
				9 & EDL & \tabincell{l}{Yuan et al.} & 52.4566 & 35.2163 & 69.6969 & 31.0286
				\tabularnewline
				10 & NJUST-MILab & \tabincell{l}{Wang et al.} & 50.7498 & 34.4634 & 67.0362 & 14.5339
				\tabularnewline
				11 & Hot Search Squad & \tabincell{l}{Jie et al.} & 48.5406 & 21.6193 & 75.6419 & 75.7045
				\tabularnewline
				\hline
		\end{tabular}}
	\end{table*}

	\begin{table*}
		\renewcommand\arraystretch{1.5}
		\caption{The award-winning results in the track of lightweight infrared small target detection.}
		\label{tab3}
		\centering
		\scriptsize
		\setlength{\tabcolsep}{1.2mm}{
			\begin{tabular}{cllccccccc}
				\hline
				Rank &Team
				& Authors
				& Score
				& mIoU ($\times 10^{-2}$)
				& $P_d$ ($\%$)
				& $F_a$ ($\times 10^{-6}$)
				& Params ($M$)
				& GFLOPs
				
				\tabularnewline
				\hline
				1 & Chainey & \tabincell{l}{Chenxu Peng} & 77.6729 & 33.8738 & 78.4534 & 60.9312 & 0.0288 & 0.0426
				\tabularnewline
				2 & Stars Twinkle and Shine & \tabincell{l}{Xiang et al.} & 77.0699 & 38.2523 & 75.3753 & 32.7550 & 0.0469 & 0.4065
				\tabularnewline 
				3 & MCV-TEAM & \tabincell{l}{Yu et al.} & 76.2318 & 30.5698 & 77.2710 & 46.4470 & 0.0199 & 0.253
				\tabularnewline
				4 & BIT-CQRS & \tabincell{l}{Huang et al.} & 75.8502 & 36.4748 & 70.7957 & 30.9488 & 0.0262 & 0.3379
				\tabularnewline
				5 & 311-IRSTD & \tabincell{l}{Wu et al.} & 75.3376 & 36.5870 & 69.3506 & 32.6290 & 0.0357 & 0.3742
				\tabularnewline
				6 & Lightweight Infrared Intelligent Detection & \tabincell{l}{Kou et al.} & 74.9502 & 33.1144 & 69.7635 & 48.4658 & 0.0075 & 0.3438
				\tabularnewline
				7 & 4L\&1H & \tabincell{l}{Li et al.} & 74.6304 & 34.3826 & 69.5758 & 40.9610 & 0.0399 & 0.4574
				\tabularnewline
				8 & CQU-TEAM & \tabincell{l}{Liu et al.} & 74.4930 & 33.8961 & 67.0983 & 42.1918 & 0.0230 & 0.2493
				\tabularnewline
				9 & MIG & \tabincell{l}{Zhou et al.} & 74.3454 & 31.3492 & 69.3130 & 64.5594 & 0.0200 & 0.2989
				\tabularnewline
				10 & NpuBugFree & \tabincell{l}{Lei et al.} & 73.5449 & 31.3861 & 67.4924 & 60.1341 & 0.0445 & 0.3388
				\tabularnewline
				11 & Daimaqiaodedui & \tabincell{l}{Peng et al.} & 73.3859 & 31.6169 & 69.1816 & 50.1419 & 0.0671 & 0.5322
				\tabularnewline
				\hline
		\end{tabular}}
	\end{table*}

		\subsection{Track 1:  Weakly Supervised Infrared Small Target Detection Under Single Point Supervision.}
    In the 46 registered teams, 15 successfully submitted their results on the test dataset. The results of award-winning teams are reported in Table \ref{tab1}.
	
	\noindent \textbf{Architectures and main ideas.}
	In the top-performing methods, it can be summarized into two strategies to generate masks for training. The first category used two-stage pipeline. The pseudo labels are generated before training by using pretrained models (e.g., RobustSAM\cite{Chainey1}) or traditional methods (e.g., MCLC\cite{Chainey2}). The second category adopted the one-stage pipeline to update labels during training (e.g., LESPS\cite{LESPS}). Moreover, some teams also considered the imbalance of targets and background, and made the network focus on target feature learning.
		
	\noindent \textbf{Conclusions.}
	1) The submitted methods improve the state-of-the-arts of weakly-supervised IRSTD.
	2) \yq{Generating high quality labels are critical to the weakly-supervised IRSTD task for improved performance.}
	3) \yq{Balancing the targets and background during training can further improve detection results.}

	\subsection{Track 2: Lightweight Infrared Small Target Detection Under Pixel-level Supervision.}
	In the 60 registered teams, 24 successfully submitted their results on the test dataset. The results of award-winning teams are reported in Table \ref{tab3}.

	\noindent \textbf{Architectures and main ideas.}
    The top one team achieved lightweight IRSTD via knowledge distillation, while other teams by designing lightweight network structure or pruning networks. Some teams used data augmentation techniques to obtain better detection performance.

	\noindent \textbf{Conclusions.}
	1) The proposed methods improve the state-of-the-art of lightweight IRSTD.
	2) \yq{Knowledge distillation, pruning, and lightweight designing can help the network lightweight and efficient while maintaining decent detection performance.}
	3) \yq{Data augmentation techniques have positive effects on detection performance.}

\section{Competition Methods and Teams}\label{sec5}
	
	In this section, the top-performing methods in each track are briefly introduced. 
	\subsection{Track 1: Weakly Supervised Infrared Small Target Detection Under Single Point Supervision}

	\subsubsection{The Chainey Team:}
    \begin{figure}
    	\centering
    	\includegraphics[width=0.5\textwidth]{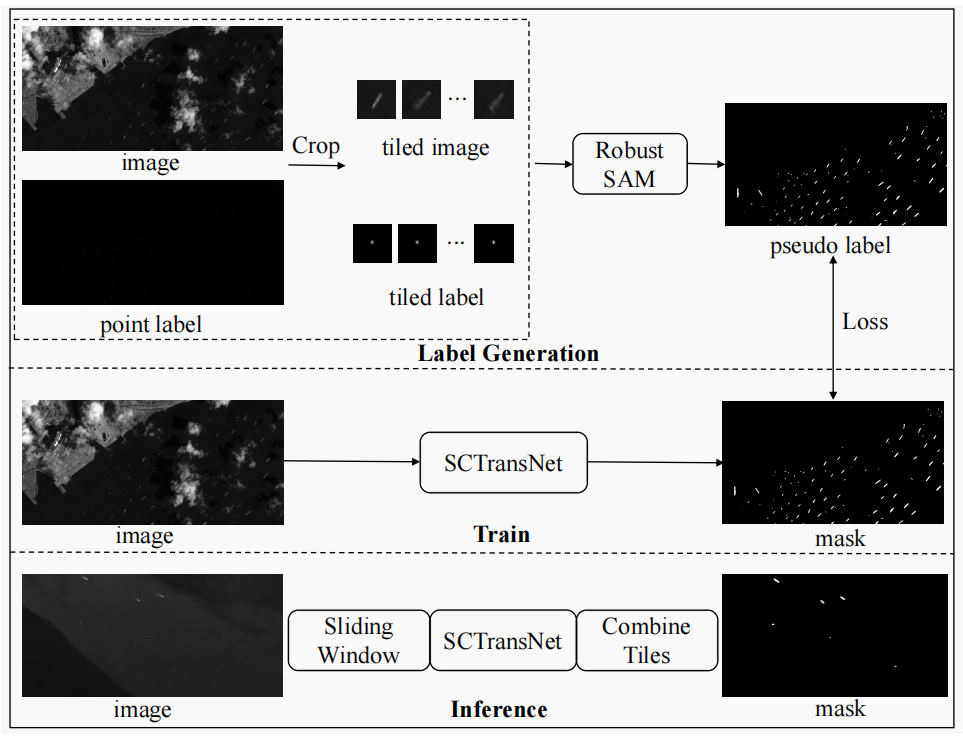}
    	\caption{An illustration of the overall framework of the Chainey Team.}
    	\label{Fig.2}
    \end{figure}
     the Chainey team ranks first in the Track 1. The objective of this task was to achieve weakly supervised infrared small target detection under single coarse point supervision (i.e., the Ground Truth (GT) point is distributed around the centroid of the GT mask following a Gaussian distribution). As shown in Fig~\ref{Fig.2}, the team employed RobustSAM \cite{Chainey1} to generate reliable pseudo labels for training the SCTransNet \cite{Stars1}. The quality of pseudo labels produced by RobustSAM is comparable to those of MCLC \cite{Chainey2} and MCGC \cite{Chainey3}, exhibiting promising performance on the SIRST and IRSTD-1K datasets\cite{Chainey4}, thus presenting itself as a potential labeling method. During the training of SCTransNet, the team adopted a stratification strategy for dataset partitioning, ensuring both class balance and uniformity of foreground pixel area, resulting in a five-fold division. For data augmentation, they utilized random cropping, rotation, flipping, and copy-paste augmentation between images of the same category within the same batch. In terms of the loss function, a deep supervision-enhanced Binary Cross-Entropy (BCE) Loss was implemented. In the inference phase, considering the substantial size of images in the test set, the team adopted a special sliding window inference strategy, selecting only the central region of the predicted images for each window. Furthermore, the team incorporated Test Time Augmentation (TTA) and Stochastic Weight Averaging (SWA) techniques. All models were realized on a single NVIDIA RTX 3090 GPU.
    	
    \subsubsection{The XJTU-IR Team:}
    \begin{figure}
    	\centering
    	\includegraphics[width=0.50\textwidth]{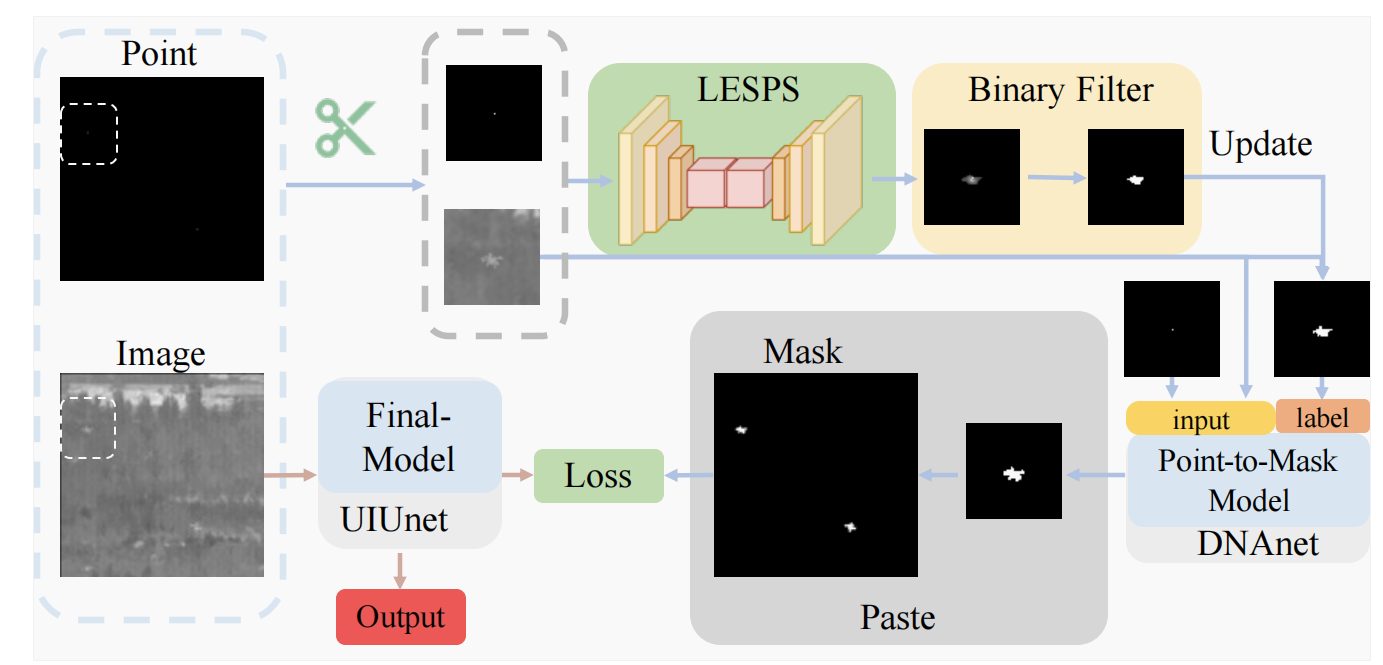}
    	\caption{An illustration of the overall framework of the XJTU-IR Team.}
    	\label{Fig.3}
    \end{figure}
    The XJTU-IR team ranks second in Track1. This team introduced an innovative Point-to-Mask strategy in the weakly supervised infrared small target detection competition under single-point supervision. This strategy divided the task into two distinct sub-tasks: (1) the high-fidelity restoration of the mask from the image-point pair, and (2) the training of the ultimate detection model utilizing the image-mask pair. A schematic representation of the entire process is depicted in the accompanying Fig.~\ref{Fig.3}.
    To mitigate the challenge of mask restoration from a single point, the team initiated the process by segmenting all images into point-centered patches. Subsequently, they employed the LESPS \cite{LESPS} algorithm to perform a preliminary mask restoration. Post binary filtering, leveraging the a priori knowledge that the target exhibits greater intensity than the background, a thresholding step then refines the detailed information within the label domain, where the mean value of the background domain is obtained through morphological dilation.
    Building on the premise that learning from simple and precise labels can enhance the network's capability to describe complex labels, the team has constructed a model based on the DNANet \cite{DNANet} architecture. This model accepted the concatenated image and point data as input, aiming to evaluate the mask's quality further. The refined mask is subsequently repositioned onto the original image based on its coordinates, serving as the basis for training the final detection model (UIU-Net \cite{UIU-Net}). In this period, the Soft-IoU Loss was adopted to replace the traditional BCE Loss. All training was implemented on 8 Tesla-100-32G GPUs.
    
    \subsubsection{The MCV-TEAM Team:}
    \begin{figure}
    	\centering
    	\includegraphics[width=0.5\textwidth]{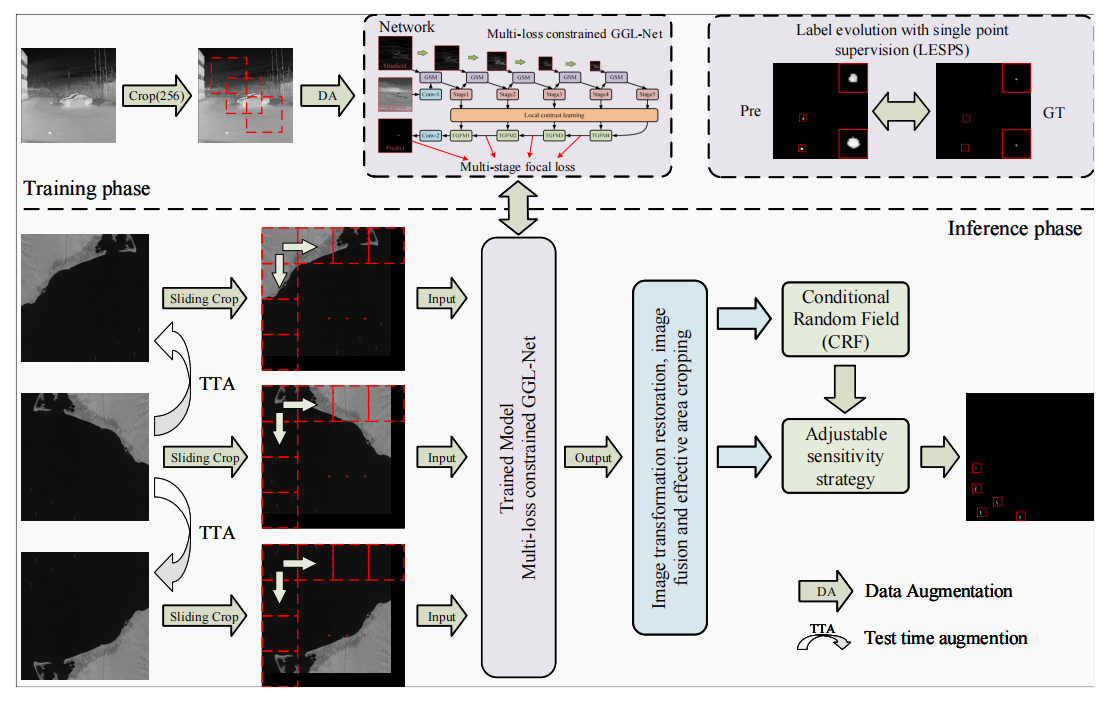}
    	\caption{An illustration of the overall framework of the MCV-TEAM.}
    	\label{Fig.4}
    \end{figure}
    the MVC-TEAM Team ranks third in Track 1. To achieve accurate detection, the team proposed a refined infrared small target detection scheme with single-point supervision \cite{MCV-TEAM1}. As shown in Fig.~\ref{Fig.4}, the scheme is divided into training phase and inference phase. During the training phase, the team introduced their previously proposed GGL-Net \cite{MCV-TEAM2} as the backbone network under the label evolution with single-point supervision (LESPS) \cite{LESPS} framework. GGL-Net is a further improved network based on MLCL-Net \cite{MCV-TEAM4} and ALCLNet \cite{MCV-TEAM5}.  GGL-Net injects multi-scale gradient magnitude images into the main branch and uses gradient information to guide network training and optimization. At the same time, the bidirectional guided fusion module (TGFM) constructed in GGL-Net can make full use of the characteristics of feature maps at different levels to effectively extract richer semantic and detail information. In addition, to further facilitate network optimization, they constructed a multi-stage loss on the original GGL-Net. During the inference phase, they built a complete post-processing strategy. Specifically, on the one hand, they adopted a combined strategy of test-time enhancement (TTA) and a conditional random field (CRF) for post-processing, which is conducive to improving the segmentation accuracy of infrared small targets. On the other hand, they introduced the adjustable sensitivity (AS) strategy \cite{MCV-TEAM7} for further post-processing. By rationally taking advantage of multiple detection results, some areas with lower confidence are added as centroid points to the finely segmented image, thereby further improving the detection rate.
    
    \subsubsection{The Stars Twinkle and Shine Team:}
    \begin{figure}
    	\centering
    	\includegraphics[width=0.5\textwidth]{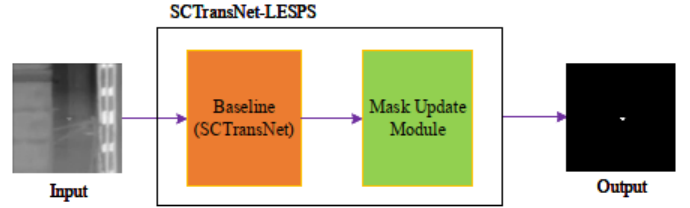}
    	\caption{An illustration of the overall framework of the Stars Twinkle and Shine Team.}
    	\label{Fig.5}
    \end{figure}
    the Stars Twinkle and Shine team ranks fourth in Track1. To achieve weakly supervised infrared small target detection under single coarse point supervision, as shown in Fig.~\ref{Fig.5}, this team proposed a solution based on spatial-channel cross transformer network (SCTransNet) \cite{Stars1} and label evolution with single point supervision (LESPS) \cite{LESPS} named SCTransNet-LESPS. SCTransNet is the baseline network of SCTransNet-LESPS to extract features of input images and predict output masks. And LESPS (mainly its mask update module) leverages the intermediate network predictions in the training phase to update the current masks, which serves as supervision until the next mask update. During training, all images were normalized and randomly cropped into patches of size 256×256 as network inputs. This team augmented the training data by random flipping, random rotation, Copy-Paste \cite{Stars3} and Mosaic \cite{Stars4}. The batch size is set to 16, and the number of training epochs is 400. Focal loss \cite{Stars5} is used to stabilize the training process. All the networks were optimized by Adam \cite{Stars6} method. Learning rate was initialized to 5e-4, and the learning rate schedule was Cosine Annealing with the hyper-parameter $T_{max}$ set to 400 and $eta_{min}$ set to 5e-6. All models were implemented in PyTorch on a PC with a Nvidia GeForce 4090 GPU.

    \subsubsection{The MIG Team:}
    \begin{figure*}
    	\centering
    	\includegraphics[width=0.99\textwidth]{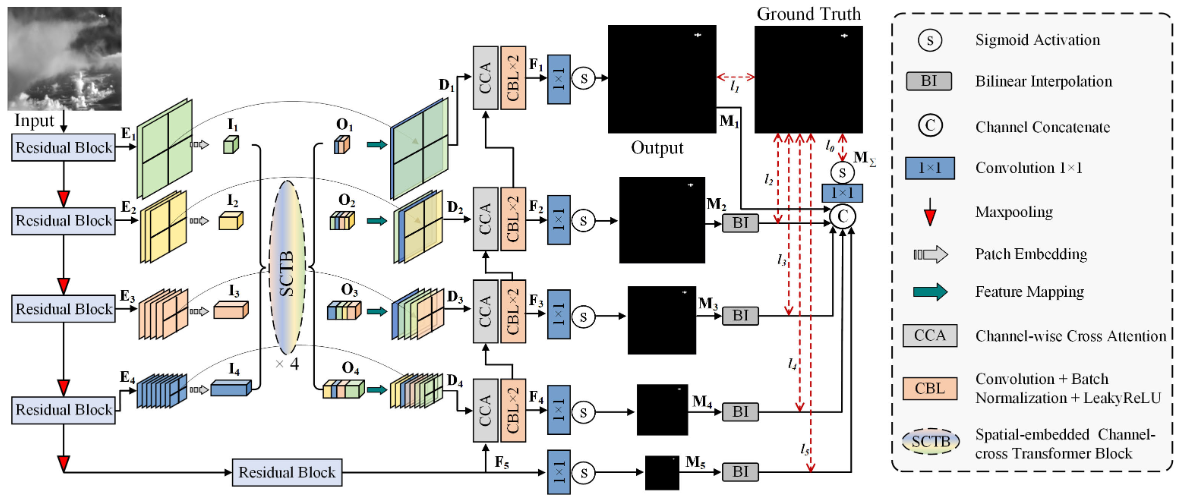}
    	\caption{An illustration of the overall framework of the MIG Team.}
    	\label{Fig.6}
    \end{figure*}
    the MIG team ranks 5th in ICPR-Track1. This team employs the weakly-supervised training framework LESPS \cite{LESPS} and chose SCTransNet \cite{Stars1} as the main model because, through comparison experiments, it worked best with LESPS among a bunch of IRSTD models. The overview of SCTransNet is shown as Fig.~\ref{Fig.6}.
    
    
    Specifically, the team observed that experiments in LESPS and SCTransNet are all conducted on three benchmark SIRST dataset (i.e., NUAA-SIRST, NUDT-SIRST and IRSTD-1K), with standard dataset spliting method. Theoretically, the combination of LESPS and SCTransNet should perform well when trained on these datasets without altering the original structure and parameters. And according to the description of the ICPR-Track1 dataset, these SIRST datasets appear to be a subset of the ICPR-Track1 dataset. Upon further examination, however, the team finds that the dataset split used in the ICPR-Track1 does not align with the standard dataset split of these SIRST datasets. As a result, the team could not directly obtain the standard SIRST training set from the ICPR-Track1 dataset. Nonetheless, taking the intersection of the ICPR-Track1 training set and these SIRST datasets as the actual training set is still a good choice. To achieve this, the team implements a simple data filtering method using the Structural Similarity Index (SSIM), resulting in a high-quality training set of 1445 images intersecting with these SIRST dataset from the 6000 images in the ICPR-Track1 training set.
    
    Through filtering, the team excluded 4555 images, including 749 land/air-based images and 3806 space-based images. Then the team conducted comparative experiments by training the same model on three different training sets: the 1445 high-quality images, 2194(749+1445) land/air-based images, and the 6000 images, and respectively evaluated the performance on the ICPR-Track1 test set. The result showed that the model trained on the 1445 high-quality images outperforms the other two models. These experiments demonstrates the effectiveness of the data filtering strategy. The high-quality training set allows the model to efficiently learn segmentation capabilities, while the reduced size of training set does not compromise the model's generalization ability in ICPR-Track1.
    
    \subsubsection{The ISTDGroup\_XDH Team:}
    \begin{figure}
    	\centering
    	\includegraphics[width=0.5\textwidth]{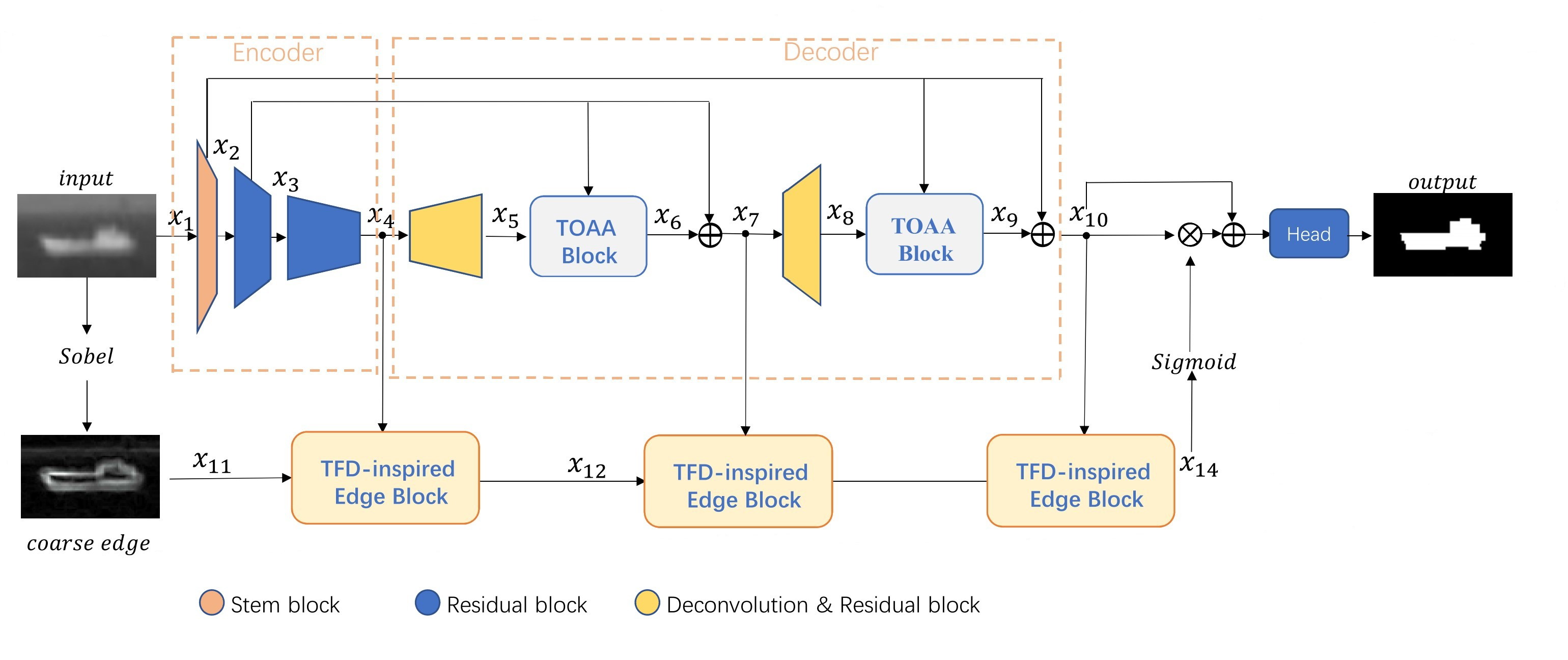}
    	\caption{An illustration of the overall framework of the ISTDGroup\_XDH Team.}
    	\label{Fig.7}
    \end{figure}
    the ISTDGroup\_XDH team ranks sixth in Track1. This team employed a label evolution framework named label evolution with single point supervision (LESPS) \cite{LESPS}, which progressively updates point labels by leveraging the intermediate outputs of the network to recover the final infrared small target masks. Specifically, LESPS progressively refines the current labels during training by leveraging the intermediate outputs of the network, using these labels as GT labels for the next label update. Through iterative label optimization, the network can eventually extract small and dim targets from complex backgrounds under single coarse point supervision, enabling pixel-level single-frame infrared small target (SIRST) detection in an end-to-end manner. Additionally, this team used ISNet \cite{ISNet} as the baseline network, where Taylor finite difference (TFD)-inspired edge block and two-orientation attention aggregation (TOAA) block are used to detect the precise shape information of infrared targets, as shown in Fig.~\ref{Fig.7}. The former block promotes crosslevel feature fusion to enhance the shape representation capacity of high-level features and the latter block extracts useful edge features to help predict accurate target mask with precise shape. During training, the input images were normalized and randomly cropped to 256×256. This team used random flipping and rotation as a data augmentation strategy. Due to the extreme positive-negative sample imbalance (less than 10 vs. more than 256×256) in SIRST detection with point supervision, they employed focal loss to stabilize the training process. The model was trained with Adam optimizer. The initial learning rate was set to 0.0005, which was reduced by ten times at the 200th and 300th epochs. The batch size was set to 24, and the number of training epochs was 800. All experiments were performed on one NVIDIA RTX 4090 GPU with 24GB memory.
    
    \subsubsection{The Banyan City Team:}
    \begin{figure}
    	\centering
    	\includegraphics[width=0.5\textwidth]{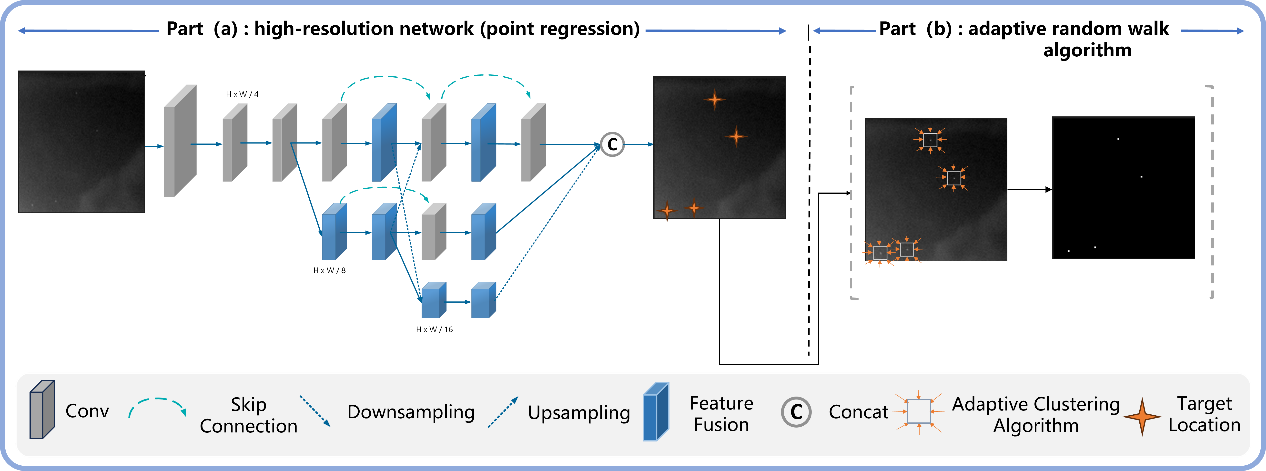}
    	\caption{An illustration of the overall framework of the Banyan City Team.}
    	\label{Fig.8}
    \end{figure}
    the Banyan City team ranks seventh in Track1. The task of infrared small target detection poses significant challenges, primarily due to two key issues: (1) The difficulty in obtaining precisely annotated data, which most existing methods rely on \cite{HJD1}; and (2) The presence of high-contrast clutter in the background, which often causes interference. To address these challenges, our team has developed a two-stage weakly supervised network for infrared small target detection (TWS-Net), as shown in Figure.\ref{Fig.8}. Firstly, they proposed a high-resolution network that utilizes point regression to acquire target location information. Subsequently, they designed an adaptive clustering algorithm based on the random walk algorithm \cite{HJD2} target contour information. During the training process, they employed the Adam optimizer and the ReduceLROnPlateau learning rate scheduler, with a learning rate reduction factor of 0.1 and a patience parameter of 3. The initial learning rate was set at 0.00005. They used a batch size of 6 and train the model for 2000 epochs. The loss function was the mean squared error, and a Gaussian kernel of size 33 was applied. The model was trained and tested using an Nvidia RTX 3090 GPU.

	\subsection{Track 2: Lightweight Infrared Small Target Detection Under Pixel-level Supervision.}

    \subsubsection{The Chainey Team:}
\begin{figure}
	\centering
	\includegraphics[width=0.45\textwidth]{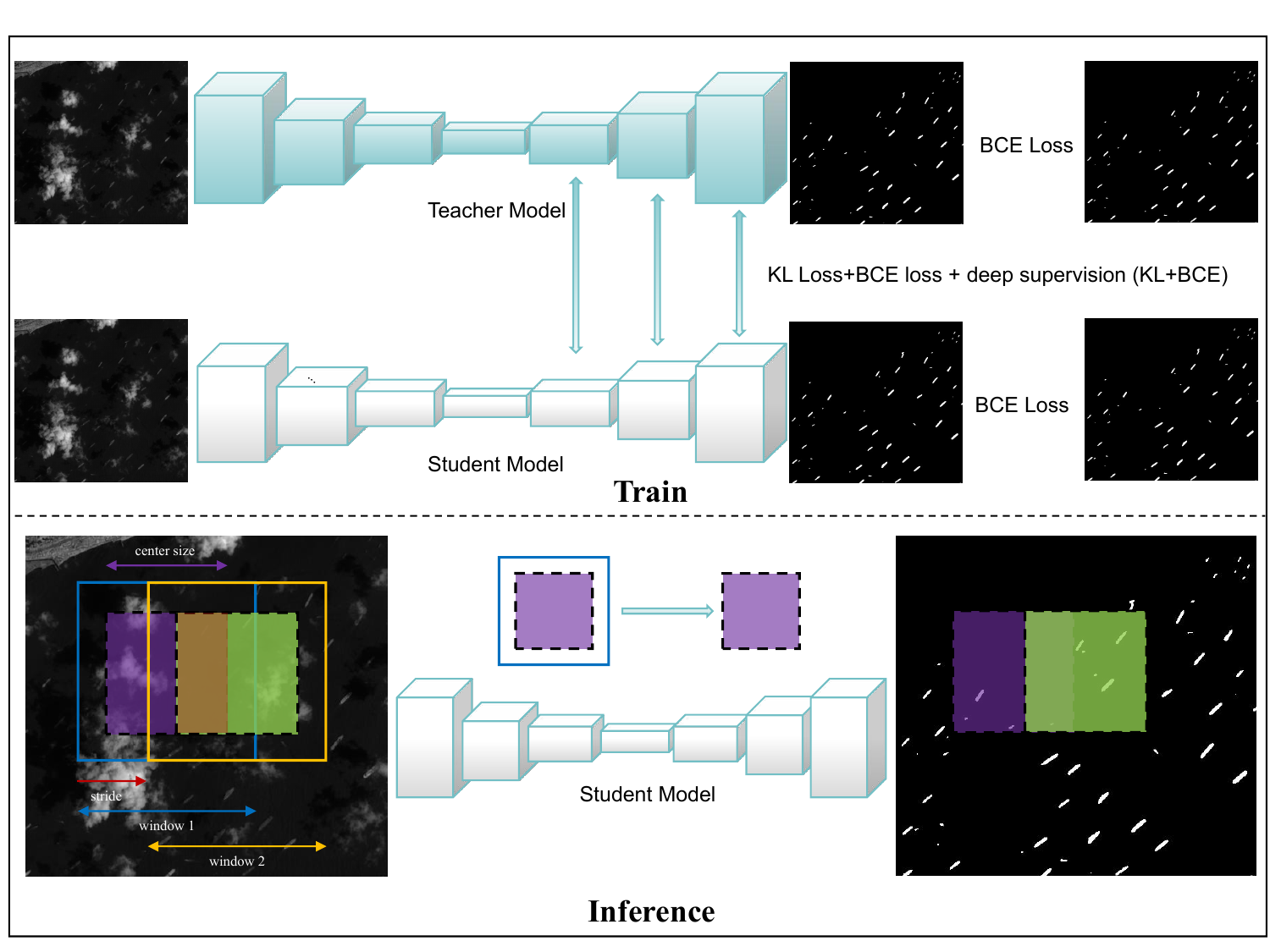}
	\caption{An illustration of the overall framework of the Chainey Team.}
	\label{Fig.9}
	\end{figure}

    \yq{the Chainey team ranks the first in the Track 2. As shown in Fig.~\ref{Fig.9}, this team employed the full SCTransNet \cite{Sctransnet} model as the teacher model and a smaller version as the student model, leveraging knowledge distillation to enhance segmentation accuracy. Initially, they trained the model without knowledge distillation and introduced the distillation loss function after the model had converged, subsequently training for more epochs than the teacher model \cite{knowledge}. In training SCTransNet, the team applied a five-fold cross-validation strategy that balanced class distribution and foreground pixel area. Data augmentation techniques included random cropping, rotation, and flipping, as well as copy-paste augmentation within the same batch for images of the same class \cite{DNANet}. The loss function utilized BCE Loss with deep supervision and KL Loss with deep supervision, with the model fine-tuned using Lovasz Loss \cite{lovasz} in the later stages. During inference, considering the large size of images in the test set, the team adopted a special sliding window inference strategy, selecting only the central region of the predicted image for each window. Additionally, they employed Test Time Augmentation (TTA) and Stochastic Weight Averaging \cite{sgd} (SWA) techniques. All models were implemented on a single NVIDIA RTX 3090 GPU.}

    \begin{figure}
   	\centering
   	\includegraphics[width=0.45\textwidth]{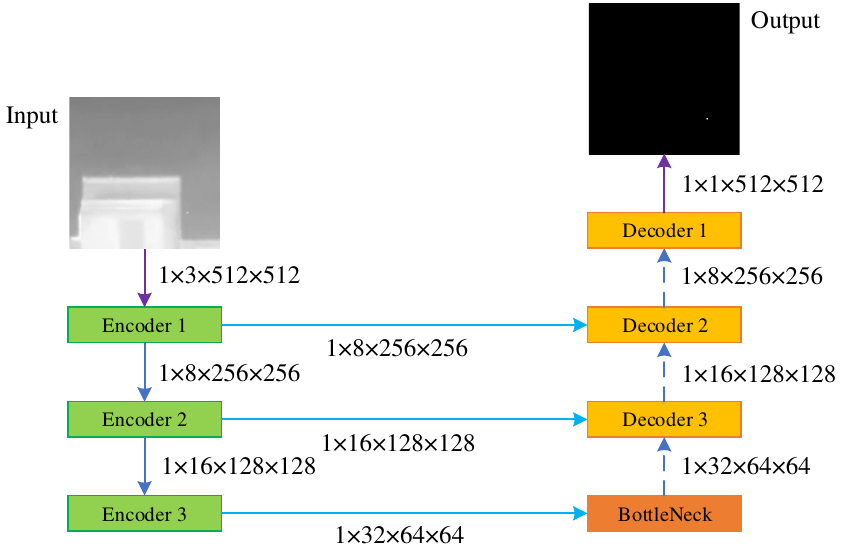}
   	\caption{An illustration of the overall framework of LWNet, which consists of 7 components: Encoder 1, Encoder 2, Encoder 3, BottleNeck,  Decoder 1, Decoder 2 and Decoder 3.}
   	\label{Fig.10}
    \end{figure}

       \begin{figure*}
   	\centering
   	\includegraphics[width=0.80\textwidth]{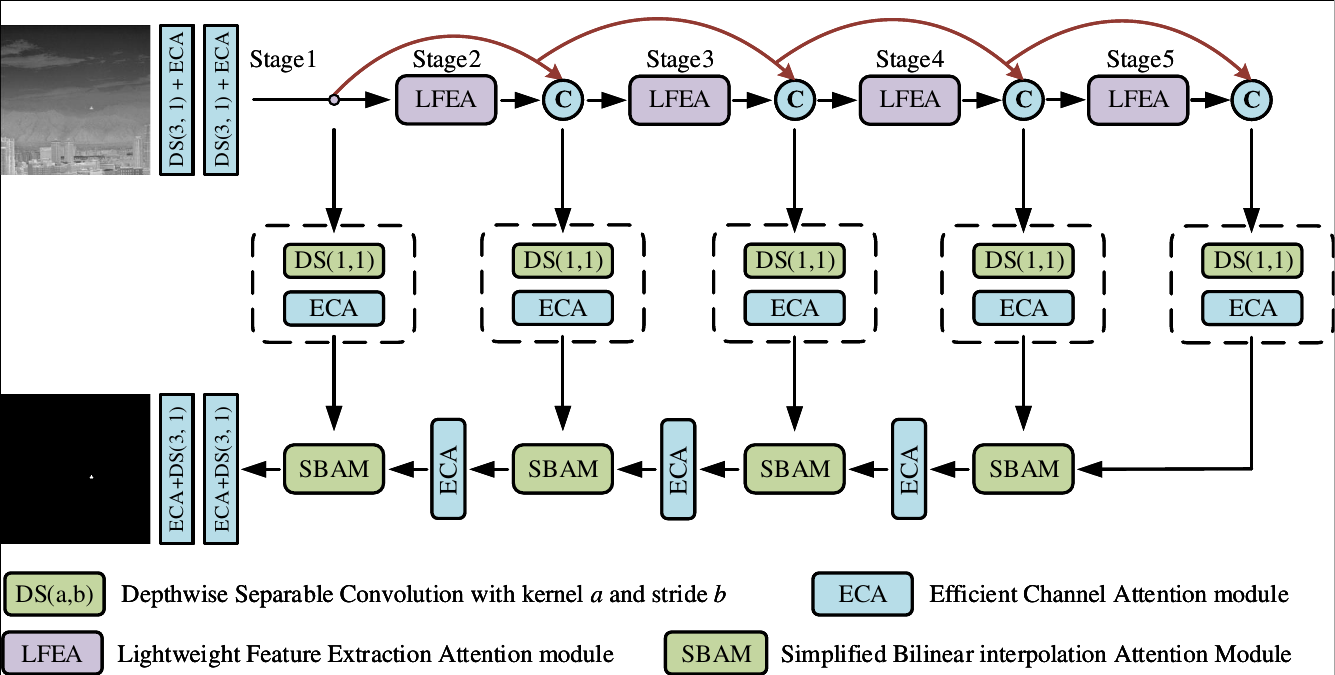}
   	\caption{An illustration of the overall framework of LR-Net.}
   	\label{Fig.11}
   \end{figure*}
	
    \subsubsection{The Stars Twinkle and Shine Team:}
    \yq{the Stars Twinkle and Shine team ranks second in Track2. To achieve lightweight and efficient infrared small target detection under pixel level supervision, this team proposed a lightweight network (LWNet), as shown in Fig.~\ref{Fig.10}. An input image was fed to the network through 3 stages: encoder stage, bottleneck stage and decoder stage. Specifically, LWNet consists of 7 components. (1) Encoder 1: Spatial and Channel Squeeze \& Excitation (SCSE) \cite{Concurrent} module was adopted to boost meaningful features, while suppressing weak ones. (2) Encoder 2: Ordinary convolution blocks and scSE blocks were used to extract features. (3) Encoder 3: Based on depthwise convolution, atrous convolution and asymmetric convolution, a module named depthwise-atrous-asymmetric-atrous (DAAA) module was proposed to extract deeper features. (4) BottleNeck: An Axial Depthwise Convolution module \cite{light2} was adopted to capture multi-spatial representations of high-level features at the bottom stage. (5) Decoder 3: First upsampled features using transposed convolution, then fused the up-sampled features with the skip-connection features from Encoder 2, and finally processed these features based on attention mechanism. (6) Decoder 2: The components of this decoder were the same as Decoder 3, but the parameters were different. (7) Decoder 1: A transposed convolution was used to upscale features to the original shape and a sigmoid function was adopted to generate output.}

    \subsubsection{The MVC-TEAM Team:}
    \yq{the MVC-TEAM Team ranks third in Track 2. The team built a lightweight and robust infrared small target detection network (LR-Net) \cite{LR-Net}, which abandons the complex structure and achieves an effective balance between detection accuracy and resource consumption. The proposed LR-Net is lightweight processing and innovation based on MLCL-Net \cite{yu2022infrared} and ALCL-Net \cite{yu2022pay}. The structure of LR-Net is shown in Fig.\ref{Fig.11}. It can be divided into three parts: feature extraction (top), feature transfer (middle) and feature fusion (bottom). In the feature extraction part, they propose a lightweight feature extraction attention (LFEA) module, which can fully extract effective features and enhance the interaction between feature channels while maintaining lightweight. At the same time, they also proposed a low-layer feature distribution (LFD) strategy, which downsamples the underlying features via max pooling and then concatenates them with the high-level feature map in the feature channel dimension, thereby greatly avoiding the loss of infrared small targets in the high-level feature map. In the feature transfer part, a layer of depth-wise separable convolution and an ECA layer are used, which helps to improve the effectiveness of information transfer while ensuring lightweight. In the feature fusion part, it introduces the simplified bilinear interpolation attention module (SBAM) proposed in ALCL-Net \cite{yu2022pay}. This module uses low-level features to constrain the attention of high-level features and fuses the low-level features with the constrained high-level features. It is worth noting that they use the test-time augmentation strategy and their proposed adjustable sensitivity (AS) strategy \cite{zhao2024infrared,MCV-TEAM1} as post-processing strategies in the inference phase. This strategy significantly improves the detection rate of infrared small targets while ensuring the detection accuracy of the model. LR-Net makes the network as lightweight as possible while ensuring detection accuracy. In summary, LR-Net has very few parameters and computational complexity and has excellent overall performance.}
   
       \begin{figure}
	\centering
	\includegraphics[width=0.5\textwidth]{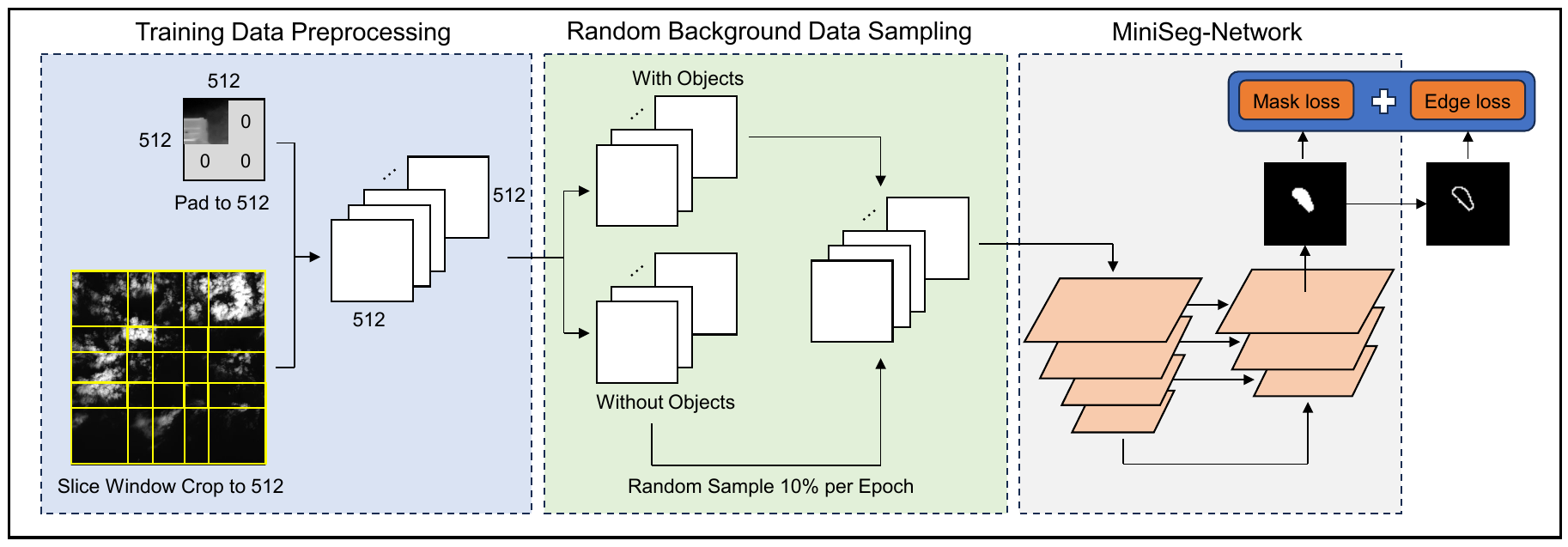}
	\caption{An illustration of the overall framework of the BIT-CQRS Team.}
	\label{Fig.12}
	\end{figure}

    \begin{figure}
	\centering
	\includegraphics[width=0.5\textwidth]{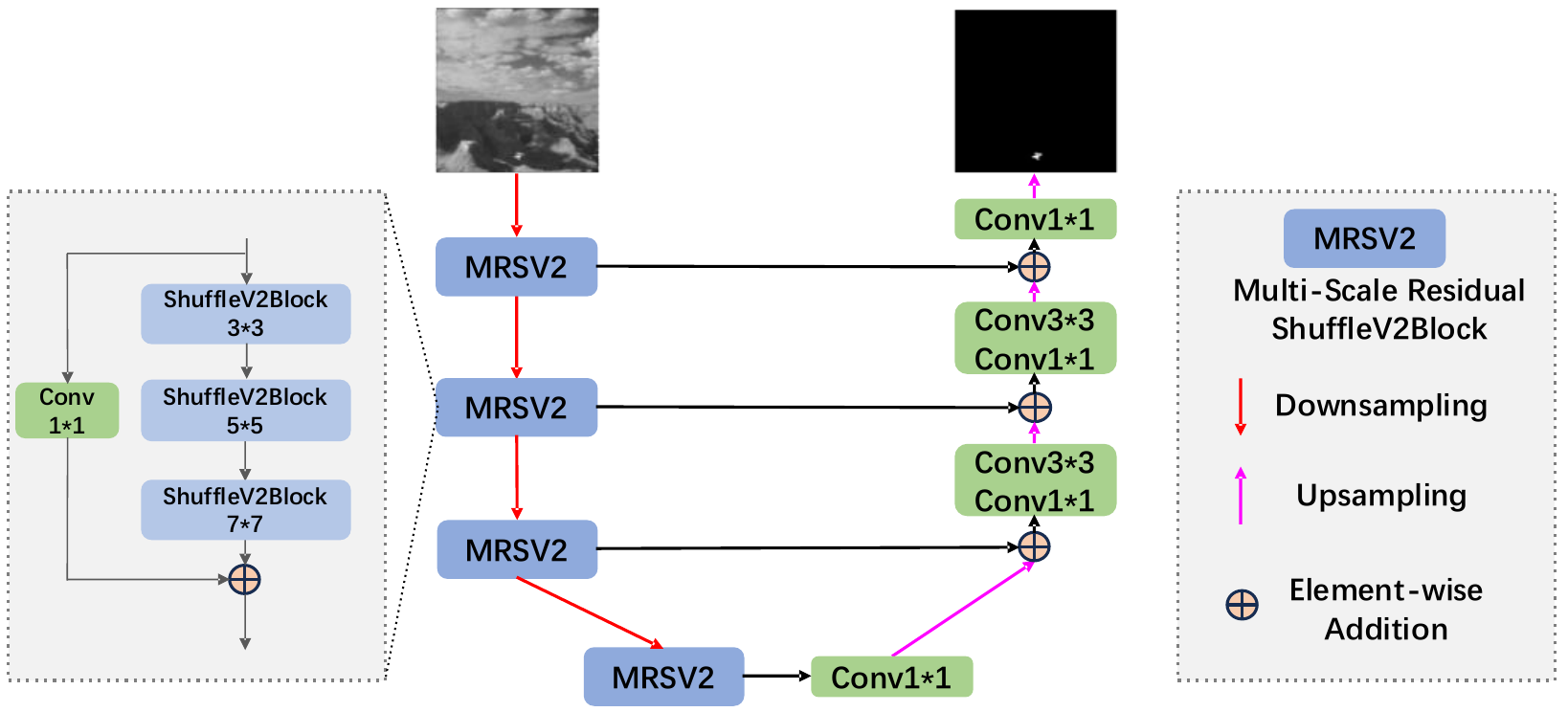}
	\caption{The overall structure of a Multi-Scale Residual Channel Shuffling Network(MRCSNet).}
	\label{Fig.13}
	\end{figure}

     \begin{figure*}
    \centering
    \includegraphics[width=0.99\textwidth]{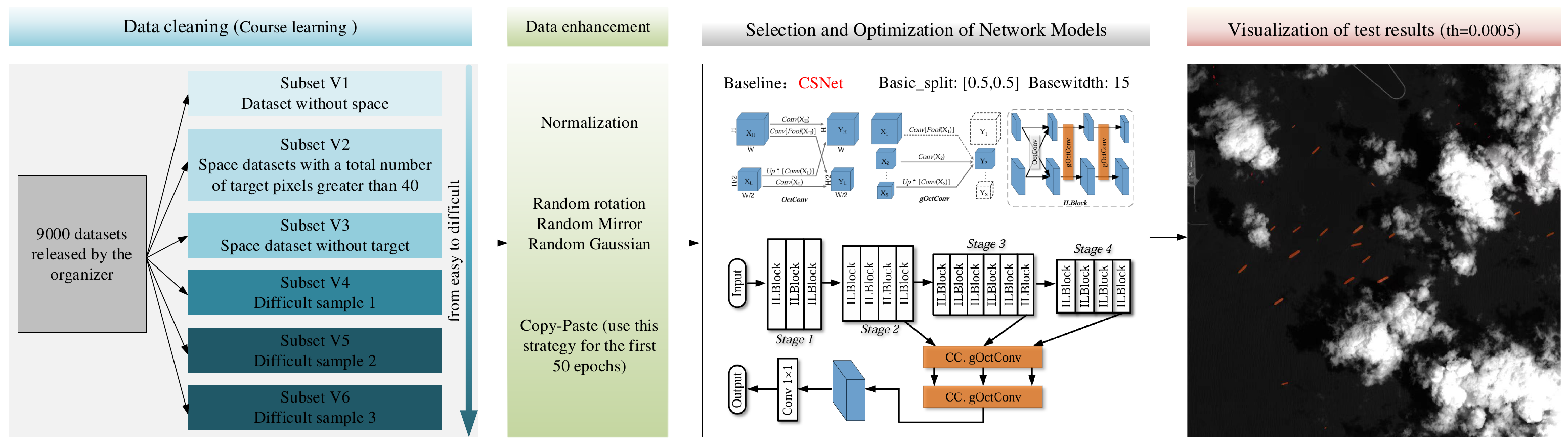}
    \caption{An illustration of the overall framework of the Lightweight infrared intelligent detection Team.}
    \label{Fig.14}
	\end{figure*}

    \begin{figure}
	\centering
	\includegraphics[width=0.5\textwidth]{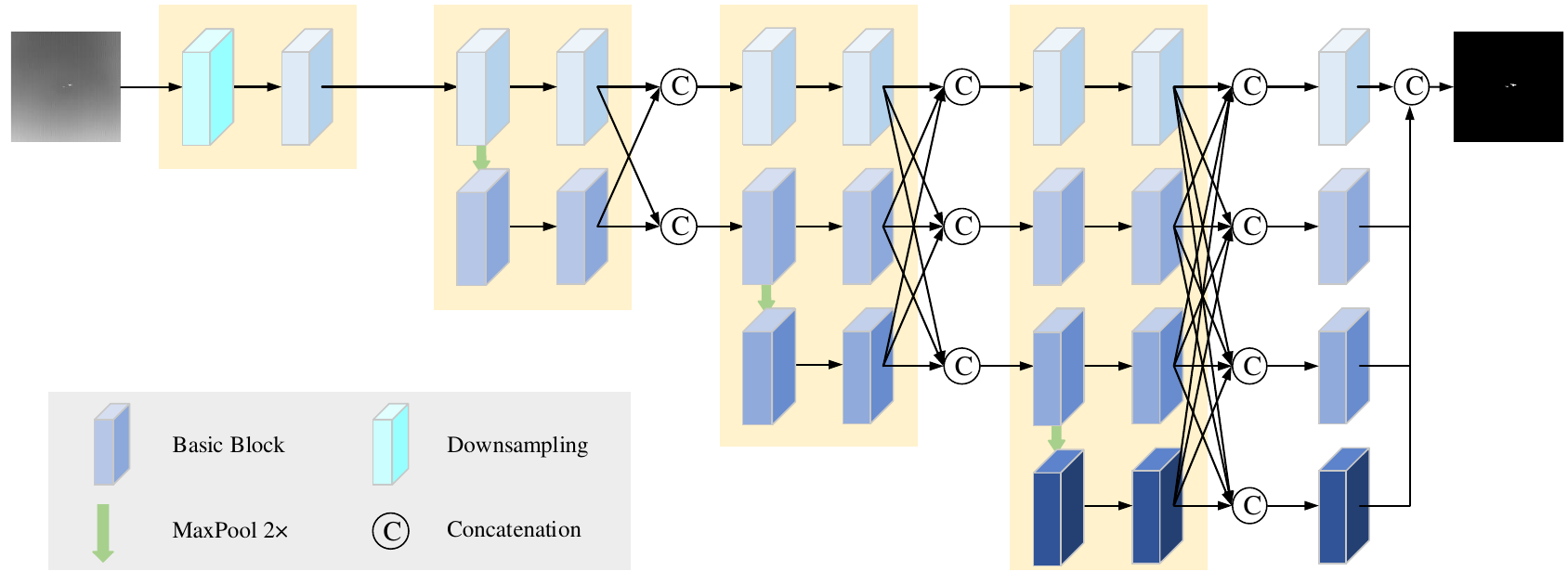}
	\caption{An illustration of the overall framework of the CQU Team.}
	\label{Fig.15}
	\end{figure}

    \subsubsection{The BIT-CQRS Team:}
	\yq{the BIT-CQRS team ranks fourth in Track2. As shown in Fig.~\ref{Fig.12}, this team used MiniSeg as the segmentation network. With a very low parameter number of 83k and high computational efficiency, MiniSeg is able to simultaneously meet the needs of high accuracy and low computational costs for infrared small object detection. To address the issue of inconsistent image size in the dataset, this team expanded images smaller than 512$\times$512 to 512$\times$512 by zero-value padding, and cropped images larger than 512$\times$512 into a number of slices with a size of 512$\times$512 by means of slice window with an overlapping of 64. In order to solve the problem of massive background slices, this team proposed the random background data sampling strategy, which resets the items in the training dataset for each epoch. Specifically, for each reset, all slices containing objects are retained, and 10\% of the background slices are randomly sampled, which together with the object slices form the training data for this epoch. Data augmentation strategies used include: mosaic, random flip, random scaling and random cropping. Training was performed using AdamW and cosine annealing with an initial learning rate of 5e-4. Batch size was set to 16 and training lasted for a total of 500 epochs. In order to further reduce the network parameters and improve the computational efficiency, this team pruned the number of channels and the number of module stacks at each level of the original model. As a result, the parameters were trimmed from 0.083M to 0.026M, and the GFLOPs has dropped from 0.533 to 0.338. Training loss consists of two parts, namely mask loss and edge loss, which results in a more accurate shape profile of objects.}

    \subsubsection{The 311-IRSTD Team:}
	\yq{the 311-IRSTD team ranks fifth in Track 2. As shown in Fig.~\ref{Fig.13}, this team proposed a Multi-scale Residual Channel Shuffle Network(MRCSNet) for effective detection of small infrared targets. The overall structure of the model is similar to FPN(Feature Pyramid Network), with the team referencing the ShuffleNetV2 \cite{shufflenet} network to design an MRSV2(Multi-scale Residual ShuffleV2) block for extracting features of small infrared targets. After simple feature fusion, the prediction map is generated. This is a compact and efficient structure. Due to the significant differences between infrared images and natural scene images, the team calculated the mean and variance of 9000 training images and used Gaussian filtering for preprocessing. During training, the Adam algorithm was used for gradient descent optimization, with an initial learning rate set to 0.001, a batch size of 16, and 1200 training epochs. Dice Loss and Soft IoU loss were used jointly to address class imbalance, represented as loss = (1-$\gamma$)Dice + $\gamma$IoU, with the hyperparameter $\gamma$ set to 0.3. The proposed model was implemented on a Nvidia A10 (24GB) GPU.}

    \subsubsection{The Lightweight infrared intelligent detection Team:}
	\yq{the Lightweight infrared intelligent detection Team ranks sixth in Track2. Their solution can both maintain the detection performance and squeeze the model size to the limit of 7.5KB, which is only 1/4 of the champion solution, far ahead of all participating teams. The scheme design process is shown in Fig.~\ref{Fig.14}, and the scheme details are as follows: (1) according to the learning strategy of the course, the 9000 datasets released by the organizer were cleaned and classified into six levels, V1 to V6, from easy to difficult; (2) They chose CSNet proposed by Professor Cheng Mingming's team at Nankai University as the baseline model; (3) based on the characteristics of infrared small targets at different scales, the network depth and width of CSNet should be designed reasonably to ensure further balance between detection accuracy and model computational complexity; (4) by coordinating multiple strategies such as data mean variance statistics (mean=[.2516, .2516, .2516], std=[.0961, .0961, .0961]), data augmentation (mirroring, flipping, Gaussian blur, copy paste), increasing input resolution (1024×1024), and reducing inference confidence threshold (0.0005), the detection accuracy of the model is further improved; (5) through a large number of ablation experiments, the optimal network structure (basic split is [0.5,0.5]; base width is 15), training hyperparameters (optimizer is SGD; initial learning rate is 0.05; learning rate reduction strategy is CosineAnnealingLR; loss function is SoftIoULoss; Epoch is 300), and training method (trained from easy to difficult according to the course learning method, using copy paste for the first 50 epochs) were obtained.}

    \subsubsection{The CQU-team:}
	\yq{the CQU-team ranks eighth in Track2. As shown in Fig.~\ref{Fig.15}, this team designed a lightweight infrared small target detection network based on the HRNet, which underwent channel pruning and was downsampled at the beginning of network. Due to the HRNet's capability to preserve the raw details of the small target, it demonstrates substantial efficacy in the domain of infrared small target detection. Following our lightweight modifications, the final model achieved an overall score of 74.493 in metrics including mIoU, Pd, Fa, Params, and GFLOPs, ranking eighth. Throughout the training phase, total 800 training epochs were employed. The initial learning rate was set at 5e-3. The optimizer "MultiStepLR" reduced the learning rate to half at 400 and 600 epochs. the loss function was computed using SoftIoULoss. All experiments were conducted on a single Nvidia RTX 4070Ti SUPER GPU.}

   \section{Conclusion}\label{sec6}
    This paper summarized the results of successful submissions to the first competition on resource-limited infrared small target detection challenge. These competition participants set new state-of-the-art for the two tracks in the field of weakly-supervised semantic segmentation and lightweight infrared small target detection. With the release of our public datasets and evaluation benchmarks, more researches in this area are expected in future.
    
   \section*{Acknowledgement}\label{sec7}
   This work was partially supported by the China Postdoctoral Science Foundation under Grant Number (No.GZB20230982 2023M744321).
    
	\section*{Teams and Affiliations}
	\label{appendix}
	
	\subsection*{The competition Organization Committee}
	\noindent \textbf{\textit{Instructors:}} Miao Li$^1$(lm8866@nudt.edu.cn), Jian Zhao$^{2,3}$, Lei Jin$^4$, Chao Xiao$^1$, Qiang Ling$^1$, Zaiping Lin$^1$, Weidong Sheng$^1$\\
	\noindent \textbf{\textit{Members:}} {Boyang Li}$^{1}$(liboyang20@nudt.edu.cn), Xinyi Ying$^1$
	Ruojing Li$^1$, Yongxian Liu$^1$, Yangsi Shi$^1$, Xin Zhang$^1$, Mingyuan Hu$^1$, Chenyang Wu$^1$, Yukai Zhang$^1$, Hui Wei$^1$, Dongli Tang$^1$
	        
	\noindent \textbf{\textit{Affiliations:}} \\
	$^1$National University of Defense Technology\\
	$^2$China Telecom Institute of AI\\
	$^3$Northwestern Polytechnical University\\
    $^4$Beijing University of Posts and Telecommunications\\
	
	\section*{Track 1: Weakly Supervised Infrared Small Target Detec-
tion Under Single Point Supervision.}
	\subsection*{(1) Chainey} 
        \noindent \textbf{\textit{Members:}} Chenxu Peng$^1$ (pcx0521@163.com)\\
        \noindent \textbf{\textit{Affiliations:}} \\
        $^1$Zhejiang SUPCON Information Co. Ltd.
	
	\subsection*{(2) XJTU-IR}
        \noindent \textbf{\textit{Members:}} Huoren Yang$^1$ (yanghuoren@stu.xjtu.edu.cn), Lingjie Liu$^1$, Ziqi Liu$^3$.\\
        \noindent \textbf{\textit{Instructors:}} Wei Ke$^3$ (wei.ke@xjtu.edu.cn), Yuhang He$^2$\\
        \noindent \textbf{\textit{Affiliations:}} \\
        $^1$School of Physics, Xi’an Jiaotong University\\
        $^2$College of Artificial Intelligence, Faculty of Electronic and Information Engineering, Xi’an Jiaotong University\\
        $^3$School of Software Engineering, Faculty of Electronic and Information Engineering, Xi’an Jiaotong University\\

        \subsection*{(3) MCV-TEAM}
        \noindent \textbf{\textit{Members:}} Chuang Yu$^{1,2,3}$ (yuchuang@sia.cn), Jinmiao Zhao$^{1,2,3}$\\
	\noindent \textbf{\textit{Instructors:}} Zelin Shi$^{1,2}$ (zlshi@sia.cn), Yunpeng Liu$^{1,2}$\\
        \noindent \textbf{\textit{Affiliations:}} \\
        $^1$The Key Laboratory of Opto-Electronic Information Processing, Chinese Academy of Sciences, Shenyang 110016, China\\
        $^2$The Shenyang Institute of Automation, Chinese Academy of Sciences, Shenyang 110016, China\\
        $^3$The School of Computer Science and Technology, University of Chinese Academy of Sciences, Beijing 100049, China\\

        \subsection*{(4) Stars Twinkle and Shine}
        \noindent \textbf{\textit{Members:}} Heng Xiang$^{1}$ (stephenxiang24@163.com), You Li$^{1}$\\
	\noindent \textbf{\textit{Instructors:}} Shan Yang$^{1}$ (yangshanbuaa@163.com)\\
        \noindent \textbf{\textit{Affiliations:}} \\
        $^1$Sichuan Zhongke Langxing Photoelectric Technology Co., Ltd.\\

        \subsection*{(5) MIG}
        \noindent \textbf{\textit{Members:}} Minghang Zhou$^1$ (zminghang@std.uestc.edu.cn), Chenxi Lan$^1$, Dongyu Xie$^1$, Chaofan Qiao$^1$, Yupeng Gao$^1$
        \noindent \textbf{\textit{Instructors:}} Guoqing Wang$^1$ (gqwang0420@uestc.edu.cn), Tianyu Li$^1$\\
        \noindent \textbf{\textit{Affiliations:}} \\
        $^1$Center for Future Media University of Electronic Science and Technology of China\\

        \subsection*{(6) ISTDGroup\_XDH}
        \noindent \textbf{\textit{Members:}} Zhihao Ma$^1$ (23021211614@stu.xidian.edu.cn), Deping Chen$^1$, Xiaopeng Song$^1$, Jiuping Yang$^1$\\
	\noindent \textbf{\textit{Instructors:}}  Yongxu Liu$^1$ (yongxu.liu@xidian.edu.cn), Wei Feng$^2$\\
        \noindent \textbf{\textit{Affiliations:}} \\
        $^1$Hangzhou Institute of Technology, Xidian University\\
        $^2$School of Electronic Engineering, Xidian University\\

        \subsection*{(7) Banyan City}
        \noindent \textbf{\textit{Members:}} Rixiang Ni$^1$ (email: 220227022@fzu.edu.cn), Changhai Luo$^1$, Ye Lin$^1$, Shuyuan Zheng$^1$.\\
        \noindent \textbf{\textit{Instructors:}} Zhaobing Qiu$^1$ (qiuzhaobing@fzu.edu.cn).\\
        \noindent \textbf{\textit{Affiliations:}} \\
        $^1$School of mechanical engineering and automation, Fuzhou University.\\
	
	\section*{Track 2: Lightweight Infrared Small Target Detection Under Pixel-level Supervision.}
	   \subsection*{(1) Chainey} 
          \noindent \textbf{\textit{Members:}} Chenxu Peng$^1$ (pcx0521@163.com)\\
          \noindent \textbf{\textit{Affiliations:}} \\
        $^1$Zhejiang SUPCON Information Co. Ltd.\\
	
        \subsection*{(2) Stars Twinkle and Shine} 
        \noindent \textbf{\textit{Members:}} Heng Xiang$^{1}$ (stephenxiang24@163.com), You Li$^{1}$\\
	\noindent \textbf{\textit{Instructors:}} Shan Yang$^{1}$ (yangshanbuaa@163.com)\\
        \noindent \textbf{\textit{Affiliations:}} \\
        $^1$Sichuan Zhongke Langxing Photoelectric Technology Co., Ltd.\\
	
    	\subsection*{(3) MCV-TEAM}
        \noindent \textbf{\textit{Members:}} Chuang Yu $^1$(yuchuang@sia.cn), Jinmiao Zhao$^1$.\\
        \noindent \textbf{\textit{Instructors:}} Yunpeng Liu$^1$(ypliu@sia.cn), Zelin Shi$^1$.\\
        \noindent \textbf{\textit{Affiliations:}} \\
        $^1$Shenyang Institute of Automation, Chinese Academy of Sciences.\\
    
    	\subsection*{(4) BIT-CQRS}
        \noindent \textbf{\textit{Members:}} Baojin Huang $^1$(272523070@qq.com), Xiaoqi Zhou$^1$, Qingshan Guo$^1$, Xuyang Zhang$^1$.\\
        \noindent \textbf{\textit{Affiliations:}} \\
        $^1$ Beijing Insitute of Technology Chongqing Innovation Center.\\

        \subsection*{(5) 311-IRSTD}
        \noindent \textbf{\textit{Members:}} Dangxuan Wu $^1$(107552204091@stu.xju.edu.cn), Jian Ma$^1$, Haodong Zeng$^1$, Luyao Wang$^1$.\\
        \noindent \textbf{\textit{Instructors:}} Xiuhong Li$^1$(xjulxh@xju.edu.cn).\\
        \noindent \textbf{\textit{Affiliations:}} \\
        $^1$ Xinjiang University.\\

        \subsection*{(6) Lightweight infrared intelligent detection}
        \noindent \textbf{\textit{Members:}} Renke Kou$^1$(krk\_Optics@aeu.edu.cn), Jian Song$^1$, Changfeng Feng$^1$, Zihao Xiong$^3$, Mengxuan Xiao$^2$\\
        \noindent \textbf{\textit{Instructors:}} Qiang Fu$^1$(Fu\_Qiang@aeu.edu.cn), Yimian Dai$^2$.\\
        \noindent \textbf{\textit{Affiliations:}} \\
        $^1$Shijiazhuang Campus, Army Engineering University of PLA.\\
        $^2$Nanjing University of Science and Technology.\\
        $^3$Henan University of Technology.\\

        \subsection*{(7) CQU-Team}
        \noindent \textbf{\textit{Members:}} Yingxu Liu$^1$(liuyingxu@stu.cqu.edu.cn), Quanyi Zhao$^1$.\\
        \noindent \textbf{\textit{Instructors:}} Hong Huang$^1$(hhuang@cqu.edu.cn).\\
        \noindent \textbf{\textit{Affiliations:}} \\
        $^1$Chongqing University.\\

\bibliographystyle{IEEEtran}
\bibliography{IEEEexample}

\begin{thebibliography}{10}
\providecommand{\url}[1]{#1}
\csname url@samestyle\endcsname
\providecommand{\newblock}{\relax}
\providecommand{\bibinfo}[2]{#2}
\providecommand{\BIBentrySTDinterwordspacing}{\spaceskip=0pt\relax}
\providecommand{\BIBentryALTinterwordstretchfactor}{4}
\providecommand{\BIBentryALTinterwordspacing}{\spaceskip=\fontdimen2\font plus
\BIBentryALTinterwordstretchfactor\fontdimen3\font minus
  \fontdimen4\font\relax}
\providecommand{\BIBforeignlanguage}[2]{{%
\expandafter\ifx\csname l@#1\endcsname\relax
\typeout{** WARNING: IEEEtran.bst: No hyphenation pattern has been}%
\typeout{** loaded for the language `#1'. Using the pattern for}%
\typeout{** the default language instead.}%
\else
\language=\csname l@#1\endcsname
\fi
#2}}
\providecommand{\BIBdecl}{\relax}
\BIBdecl

\bibitem{li2024mixed}
B.~Li, L.~Wang, Y.~Wang, T.~Wu, Z.~Lin, M.~Li, W.~An, and Y.~Guo,
  ``Mixed-precision network quantization for infrared small target
  segmentation,'' \emph{IEEE Transactions on Geoscience and Remote Sensing},
  2024.

\bibitem{2022light}
Z.~Lin, B.~Li, M.~Li, L.~Wang, T.~Wu, Y.~Luo, C.~Xiao, R.~Li, and W.~An,
  ``Light-weight infrared small target detection combining cross-scale feature
  fusion with bottleneck attention module,'' \emph{Journal of Infrared and
  Millimeter Waves}, vol.~41, no.~6, pp. 1102--1112, 2022.

\bibitem{liu2021nonconvex}
T.~Liu, J.~Yang, B.~Li, C.~Xiao, Y.~Sun, Y.~Wang, and W.~An, ``Nonconvex tensor
  low-rank approximation for infrared small target detection,'' \emph{IEEE
  Transactions on Geoscience and Remote Sensing}, vol.~60, pp. 1--18, 2021.

\bibitem{wu2023mtu}
T.~Wu, B.~Li, Y.~Luo, Y.~Wang, C.~Xiao, T.~Liu, J.~Yang, W.~An, and Y.~Guo,
  ``Mtu-net: Multilevel transunet for space-based infrared tiny ship
  detection,'' \emph{IEEE Transactions on Geoscience and Remote Sensing},
  vol.~61, pp. 1--15, 2023.

\bibitem{li2023direction}
R.~Li, W.~An, C.~Xiao, B.~Li, Y.~Wang, M.~Li, and Y.~Guo, ``Direction-coded
  temporal u-shape module for multiframe infrared small target detection,''
  \emph{IEEE Transactions on Neural Networks and Learning Systems}, 2023.

\bibitem{liu2023infrared}
T.~Liu, J.~Yang, B.~Li, Y.~Wang, and W.~An, ``Infrared small target detection
  via nonconvex tensor tucker decomposition with factor prior,'' \emph{IEEE
  Transactions on Geoscience and Remote Sensing}, 2023.

\bibitem{NUAA-SIRST}
Y.~Dai, X.~Li, F.~Zhou, Y.~Qian, Y.~Chen, and J.~Yang, ``One-stage cascade
  refinement networks for infrared small target detection,'' \emph{IEEE
  Transactions on Geoscience and Remote Sensing}, vol.~61, pp. 1--17, 2023.

\bibitem{ISNet}
M.~Zhang, R.~Zhang, Y.~Yang, H.~Bai, J.~Zhang, and J.~Guo, ``Isnet: Shape
  matters for infrared small target detection,'' in \emph{Proceedings of the
  IEEE/CVF Conference on Computer Vision and Pattern Recognition}, 2022, pp.
  877--886.

\bibitem{IRDST}
H.~Sun, J.~Bai, F.~Yang, and X.~Bai, ``Receptive-field and direction induced
  attention network for infrared dim small target detection with a large-scale
  dataset irdst,'' \emph{IEEE Transactions on Geoscience and Remote Sensing},
  vol.~61, pp. 1--13, 2023.

\bibitem{DNANet}
B.~Li, C.~Xiao, L.~Wang, Y.~Wang, Z.~Lin, M.~Li, W.~An, and Y.~Guo, ``Dense
  nested attention network for infrared small target detection,'' \emph{IEEE
  Transactions on Image Processing}, vol.~32, pp. 1745--1758, 2022.

\bibitem{NUDT-SIRST-Sea}
T.~Wu, B.~Li, Y.~Luo, Y.~Wang, C.~Xiao, T.~Liu, J.~Yang, W.~An, and Y.~Guo,
  ``Mtu-net: Multilevel transunet for space-based infrared tiny ship
  detection,'' \emph{IEEE Transactions on Geoscience and Remote Sensing},
  vol.~61, pp. 1--15, 2023.

\bibitem{DTUM}
R.~Li, W.~An, C.~Xiao, B.~Li, Y.~Wang, M.~Li, and Y.~Guo, ``Direction-coded
  temporal u-shape module for multiframe infrared small target detection,''
  \emph{IEEE Transactions on Neural Networks and Learning Systems}, 2023.

\bibitem{Anti-UAV}
N.~Jiang, K.~Wang, X.~Peng, X.~Yu, Q.~Wang, J.~Xing, G.~Li, G.~Guo, Q.~Ye,
  J.~Jiao \emph{et~al.}, ``Anti-uav: a large-scale benchmark for vision-based
  uav tracking,'' \emph{IEEE Transactions on Multimedia}, vol.~25, pp.
  486--500, 2021.

\bibitem{Chainey1}
W.-T. Chen, Y.-J. Vong, S.-Y. Kuo, S.~Ma, and J.~Wang, ``Robustsam: Segment
  anything robustly on degraded images,'' in \emph{Proceedings of the IEEE/CVF
  Conference on Computer Vision and Pattern Recognition (CVPR)}, June 2024, pp.
  4081--4091.

\bibitem{Chainey2}
B.~Li, Y.~Wang, L.~Wang, F.~Zhang, T.~Liu, Z.~Lin, W.~An, and Y.~Guo, ``Monte
  carlo linear clustering with single-point supervision is enough for infrared
  small target detection,'' in \emph{Proceedings of the IEEE/CVF International
  Conference on Computer Vision (ICCV)}, October 2023, pp. 1009--1019.

\bibitem{LESPS}
X.~Ying, L.~Liu, Y.~Wang, R.~Li, N.~Chen, Z.~Lin, W.~Sheng, and S.~Zhou,
  ``Mapping degeneration meets label evolution: Learning infrared small target
  detection with single point supervision,'' in \emph{Proceedings of the
  IEEE/CVF Conference on Computer Vision and Pattern Recognition}, 2023, pp.
  15\,528--15\,538.

\bibitem{Stars1}
S.~Yuan, H.~Qin, X.~Yan, N.~Akhtar, and A.~Mian, ``Sctransnet: Spatial-channel
  cross transformer network for infrared small target detection,'' \emph{IEEE
  Transactions on Geoscience and Remote Sensing}, 2024.

\bibitem{Chainey3}
R.~Kou, C.~Wang, Q.~Fu, Z.~Li, Y.~Luo, B.~Li, W.~Li, and Z.~Peng, ``Mcgc: A
  multi-scale chain growth clustering algorithm for generating infrared small
  target mask under single-point supervision,'' \emph{IEEE Transactions on
  Geoscience and Remote Sensing}, 2024.

\bibitem{Chainey4}
S.~Yuan, H.~Qin, R.~Kou, X.~Yan, Z.~Li, C.~Peng, and A.-K. Seghouane, ``Beyond
  full label: Single-point prompt for infrared small target label generation,''
  \emph{arXiv preprint arXiv:2408.08191}, 2024.

\bibitem{UIU-Net}
X.~Wu, D.~Hong, and J.~Chanussot, ``Uiu-net: U-net in u-net for infrared small
  object detection,'' \emph{IEEE Transactions on Image Processing}, vol.~32,
  pp. 364--376, 2022.

\bibitem{MCV-TEAM1}
J.~Zhao, Z.~Shi, C.~Yu, and Y.~Liu, ``Refined infrared small target detection
  scheme with single-point supervision,'' \emph{arXiv preprint
  arXiv:2408.02773}, 2024.

\bibitem{MCV-TEAM2}
J.~Zhao, C.~Yu, Z.~Shi, Y.~Liu, and Y.~Zhang, ``Gradient-guided learning
  network for infrared small target detection,'' \emph{IEEE Geoscience and
  Remote Sensing Letters}, 2023.

\bibitem{MCV-TEAM4}
C.~Yu, Y.~Liu, S.~Wu, Z.~Hu, X.~Xia, D.~Lan, and X.~Liu, ``Infrared small
  target detection based on multiscale local contrast learning networks,''
  \emph{Infrared Physics \& Technology}, vol. 123, p. 104107, 2022.

\bibitem{MCV-TEAM5}
C.~Yu, Y.~Liu, S.~Wu, X.~Xia, Z.~Hu, D.~Lan, and X.~Liu, ``Pay attention to
  local contrast learning networks for infrared small target detection,''
  \emph{IEEE Geoscience and Remote Sensing Letters}, vol.~19, pp. 1--5, 2022.

\bibitem{MCV-TEAM7}
J.~Zhao, Z.~Shi, C.~Yu, and Y.~Liu, ``Infrared small target detection based on
  adjustable sensitivity strategy and multi-scale fusion,'' \emph{arXiv
  preprint arXiv:2407.20090}, 2024.

\bibitem{Stars3}
G.~Ghiasi, Y.~Cui, A.~Srinivas, R.~Qian, T.-Y. Lin, E.~D. Cubuk, Q.~V. Le, and
  B.~Zoph, ``Simple copy-paste is a strong data augmentation method for
  instance segmentation,'' in \emph{Proceedings of the IEEE/CVF conference on
  computer vision and pattern recognition}, 2021, pp. 2918--2928.

\bibitem{Stars4}
A.~Bochkovskiy, ``Yolov4: Optimal speed and accuracy of object detection,''
  \emph{arXiv preprint arXiv:2004.10934}, 2020.

\bibitem{Stars5}
T.-Y. Lin, P.~Goyal, R.~Girshick, K.~He, and P.~Doll{\'a}r, ``Focal loss for
  dense object detection,'' in \emph{Proceedings of the IEEE international
  conference on computer vision}, 2017, pp. 2980--2988.

\bibitem{Stars6}
D.~P. Kingma, ``Adam: A method for stochastic optimization,'' \emph{arXiv
  preprint arXiv:1412.6980}, 2014.

\bibitem{HJD1}
J.~Wu, R.~Ni, F.~Huang, Z.~Qiu, L.~Chen, C.~Luo, Y.~Li, and Y.~Li,
  ``Single-point supervised high-resolution dynamic network for infrared small
  target detection,'' \emph{arXiv preprint arXiv:2408.01976}, 2024.

\bibitem{HJD2}
Y.~Qin, L.~Bruzzone, C.~Gao, and B.~Li, ``Infrared small target detection based
  on facet kernel and random walker,'' \emph{IEEE Transactions on Geoscience
  and Remote Sensing}, vol.~57, no.~9, pp. 7104--7118, 2019.

\bibitem{Sctransnet}
S.~Yuan, H.~Qin, X.~Yan, N.~Akhtar, and A.~Mian, ``Sctransnet: Spatial-channel
  cross transformer network for infrared small target detection,'' \emph{IEEE
  Transactions on Geoscience and Remote Sensing}, 2024.

\bibitem{knowledge}
L.~Beyer, X.~Zhai, A.~Royer, L.~Markeeva, R.~Anil, and A.~Kolesnikov,
  ``Knowledge distillation: A good teacher is patient and consistent,'' in
  \emph{Proceedings of the IEEE/CVF conference on computer vision and pattern
  recognition}, 2022, pp. 10\,925--10\,934.

\bibitem{lovasz}
M.~Berman, A.~R. Triki, and M.~B. Blaschko, ``The lov{\'a}sz-softmax loss: A
  tractable surrogate for the optimization of the intersection-over-union
  measure in neural networks,'' in \emph{Proceedings of the IEEE conference on
  computer vision and pattern recognition}, 2018, pp. 4413--4421.

\bibitem{sgd}
P.~Izmailov, D.~Podoprikhin, T.~Garipov, D.~Vetrov, and A.~G. Wilson,
  ``Averaging weights leads to wider optima and better generalization,''
  \emph{arXiv preprint arXiv:1803.05407}, 2018.

\bibitem{Concurrent}
A.~G. Roy, N.~Navab, and C.~Wachinger, ``Concurrent spatial and channel
  ‘squeeze \& excitation’in fully convolutional networks,'' in
  \emph{Medical Image Computing and Computer Assisted Intervention--MICCAI
  2018: 21st International Conference, Granada, Spain, September 16-20, 2018,
  Proceedings, Part I}.\hskip 1em plus 0.5em minus 0.4em\relax Springer, 2018,
  pp. 421--429.

\bibitem{light2}
B.-D. Dinh, T.-T. Nguyen, T.-T. Tran, and V.-T. Pham, ``1m parameters are
  enough? a lightweight cnn-based model for medical image segmentation,'' in
  \emph{2023 Asia Pacific Signal and Information Processing Association Annual
  Summit and Conference (APSIPA ASC)}.\hskip 1em plus 0.5em minus 0.4em\relax
  IEEE, 2023, pp. 1279--1284.

\bibitem{LR-Net}
C.~Yu, Y.~Liu, J.~Zhao, and Z.~Shi, ``Lr-net: A lightweight and robust network
  for infrared small target detection,'' \emph{arXiv preprint
  arXiv:2408.02780}, 2024.

\bibitem{yu2022infrared}
C.~Yu, Y.~Liu, S.~Wu, Z.~Hu, X.~Xia, D.~Lan, and X.~Liu, ``Infrared small
  target detection based on multiscale local contrast learning networks,''
  \emph{Infrared Physics \& Technology}, vol. 123, p. 104107, 2022.

\bibitem{yu2022pay}
C.~Yu, Y.~Liu, S.~Wu, X.~Xia, Z.~Hu, D.~Lan, and X.~Liu, ``Pay attention to
  local contrast learning networks for infrared small target detection,''
  \emph{IEEE Geoscience and Remote Sensing Letters}, vol.~19, pp. 1--5, 2022.

\bibitem{zhao2024infrared}
J.~Zhao, Z.~Shi, C.~Yu, and Y.~Liu, ``Infrared small target detection based on
  adjustable sensitivity strategy and multi-scale fusion,'' \emph{arXiv
  preprint arXiv:2407.20090}, 2024.

\bibitem{shufflenet}
N.~Ma, X.~Zhang, H.-T. Zheng, and J.~Sun, ``Shufflenet v2: Practical guidelines
  for efficient cnn architecture design,'' in \emph{Proceedings of the European
  conference on computer vision (ECCV)}, 2018, pp. 116--131.

\end{thebibliography}
\end{document}